\documentclass{article}

\usepackage[preprint,nonatbib]{neurips_2021}

\usepackage[numbers,sort&compress]{natbib}

\usepackage[utf8]{inputenc} 
\usepackage[T1]{fontenc}    
\usepackage[colorlinks,citecolor=blue]{hyperref}       
\usepackage{url}            
\usepackage{booktabs}       
\usepackage{amsfonts}       
\usepackage{nicefrac}       
\usepackage{microtype}      
\usepackage{xcolor}         
\usepackage{natbib}
\usepackage{graphicx}

\title{Adaptive Latent Space Tuning for Non-Stationary Distributions}

\author{%
  Alexander Scheinker\thanks{Correspondence: \texttt{ascheink@lanl.gov}.} \\
  Accelerator Operations and Technology\\
  Los Alamos National Laboratory\\
  \And
  Frederick Cropp \\
  Department of Physics and Astronomy \\
  University of California Los Angeles \\
  \And
  Sergio Paiagua \\
  Advanced Light Source Accelerator Physics \\
  Lawrence Berkeley National Laboratory \\
  \And
  Daniele Filippetto \\
  Advanced Light Source Accelerator Physics \\
  Lawrence Berkeley National Laboratory \\
}

\begin{document}

\maketitle

\begin{abstract}
Encoder-decoder style deep convolutional neural networks (CNN) are able to extract features directly from images, mix them with scalar inputs within a general low-dimensional latent space, and generate outputs which represent complex physical phenomenon. One challenge faced by deep learning methods is modeling large non-stationary systems whose characteristics change quickly with time for which re-training is not feasible. In this paper we present a method for adaptive tuning of the low-dimensional latent space of deep encoder-decoder style CNNs based on real-time feedback to compensate for unknown and fast distribution shifts. We demonstrate the approach for predicting the properties of a time-varying charged particle beam in a particle accelerator whose initial distribution and components (accelerating electric fields and focusing magnetic fields) are quickly changing with time and may not be measurable during operations. Our method utilizes the low-dimensional latent space basis directly to generate new outputs and therefore does not require access to new input beam distributions for re-training, which is important for large systems such as particle accelerators where input distribution measurements interrupt normal operations.
\end{abstract}

\section{Introduction}
Powerful machine learning (ML)-based tools are now being used in almost all fields of scientific research \cite{hastie2009elements,murphy2012machine,lecun2015deep,jordan2015machine}. The growth of ML is due to a combination of the development of advanced algorithms such as convolutional neural networks (CNN) \cite{krizhevsky2012imagenet}, spatial transformer networks \cite{jaderberg2015spatial,pfeiffer2020adapterhub,touvron2021going}, recurrent neural networks \cite{bengio1994learning, hochreiter1997long}, analytical studies which have improved the understanding of deep learning characteristics \cite{lee2017deep,xiao2020disentangling}, deep reinforcement learning \cite{mnih2015human,sutton2018reinforcement}, the availability of affordable high performance computing (graphics processing units), and powerful open source ML packages such as TensorFlow \cite{abadi2016tensorflow,abadi2016tensorflow_2}, Caffe \cite{jia2014caffe}, CNTK \cite{seide2016cntk}, Theano \cite{team2016theano}, and PyTorch \cite{paszke2019pytorch}.

\subsection{ML Applications}
ML applications for physical systems include molecular and materials science studies \cite{butler2018machine}, for use in optical communications and photonics \cite{zibar2017machine}, for studying glassy systems \cite{cubuk2015identifying,schoenholz2016structural,schoenholz2018combining}, to accurately predict battery life in \cite{berecibar2019machine}, to accelerate lattice Monte Carlo simulations using neural networks \cite{ref_ML_Quantum_Monte_Carlo}, for studying complex networks \cite{muscoloni2017machine}, for characterizing surface microstructure of complex materials \cite{lansford2020infrared}, for chemical discovery \cite{tkatchenko2020machine}, for active matter analysis by using deep neural networks to track objects \cite{cichos2020machine}, for particle physics \cite{radovic2018machine}, for antimicrobial studies \cite{ragno2020essential}, for pattern recognition for optical microscopy images of metallurgical micro-structures \cite{bulgarevich2018pattern}, for learning Perovskit bandgaps \cite{pilania2016machine}, for real-time mapping of electron diffraction patterns to crystal orientations \cite{shen2019convolutional}, for speeding up simulations \cite{edelen2020machine}, for Bayesian optimization of free electron lasers (FEL) \cite{duris2020bayesian}, for temporal power reconstruction of FELs \cite{ren2020temporal}, for various applications at the Large Hadron Collider at CERN including optics corrections and detecting faulty beam position monitors \cite{fol2019optics,fol2019unsupervised,arpaia2021machine}, for reconstruction of a storage ring's linear optics based on Bayesian inference \cite{hao2019reconstruction}, to analyze beam position monitor placement in accelerators to find arrangements with the lowest probable predictive errors based on Bayesian Gaussian regression \cite{li2019analysis}, surrogate models that map accelerator parameters to beam properties \cite{emma2018machine,hanuka2021accurate}, for temporal shaping of electron bunches in particle accelerators \cite{wan2021machine}, for stabilization of source properties in synchrotron light sources \cite{leemann2019demonstration}, ML has been implemented with differentiable programming for physics applications \cite{schoenholz2020jax}, autoencoders have been developed to model accelerator sections \cite{zhu2021deep}, ML has been used to represent many-body interactions with restricted-Boltzmann-machine neural networks \cite{ref_ML_exact_many_body}, and a powerful approach to virtual diagnostics for particle accelerators has recently been developed utilizing ensembles of quantile regression neural networks \cite{convery2021uncertainty}.

\subsection{Non-stationary Distributions}
ML for non-stationary systems is an active field of research with recent results including the use of recurrent neural networks for speech perception in non-stationary noise \cite{goehring2019using}, a method based on convolutional and recurrent neural networks has been proposed for modeling time-varying audio processors \cite{ramirez2019general}, complex-valued neural networks have been proposed for ML on non-stationary physical data and have shown that including phase information in feature maps improves both training and inference from deterministic physical data \cite{javadi2020complex,dramsch2021complex}, and methods have been developed for non-stationary systems which detect significant changes after which the weights of neural networks are updated/re-trained with new information or are continuously trained to keep up with continuous changes \cite{calandra2012learning,koesdwiady2018non,kurle2019continual}. A powerful class of approaches has been developed for the case of covariate shift, where the input distribution $P(\mathbf{x})$ is different for training and test data, but the conditional distribution of output values $P(y|\mathbf{x})$ remains unchanged \cite{shimodaira2000improving}, based on importance-weighting (IW) techniques \cite{fishman2013monte}. Accurate density estimation is difficult, especially for high dimensional systems, and IW methods have been developed which do not require accurate density estimates using kernel mean matching methods \cite{huang2006correcting} and by minimizing the Kullback-Leibler divergence between a test data density distribution and its estimate \cite{sugiyama2007direct,sugiyama2012machine}. Novel methods have also been developed for extracting instantaneous frequencies and amplitudes in quickly changing non-stationary time-series data \cite{shea2021extraction}.

\begin{figure}
  \centering
  \includegraphics[width=0.75\textwidth]{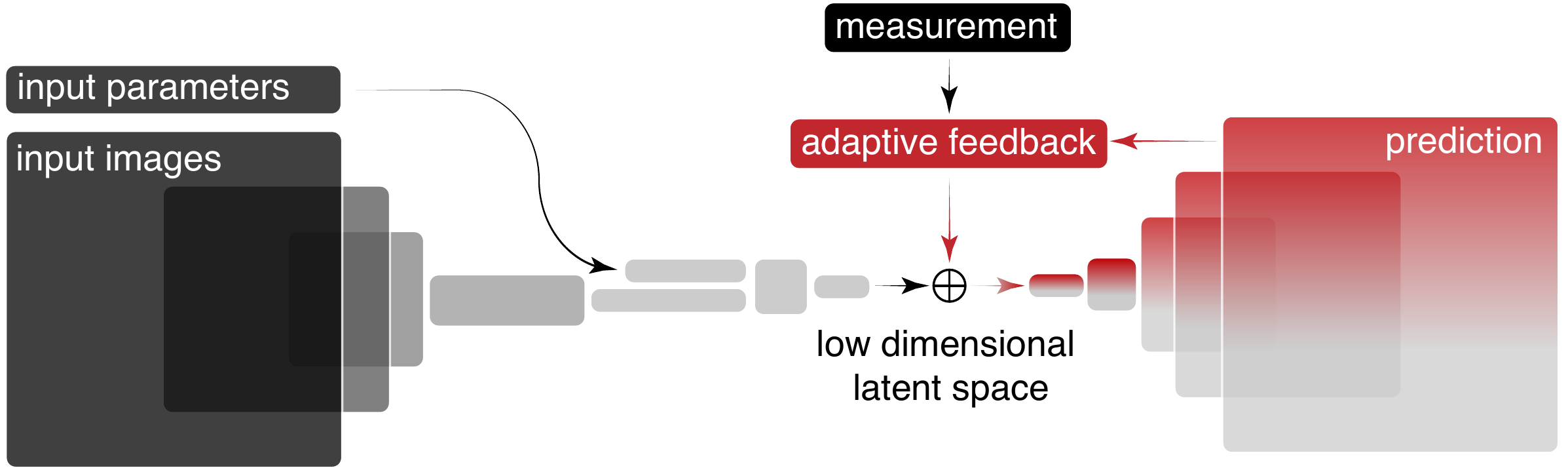}
  \caption{Adaptive machine learning setup for an encoder-decoder style 2D convolutional neural network with adaptive adjustment of the low dimensional latent space. The size of an input image is first reduced using 2D convolution layers before being flattened and concatenated with an input vector of input parameters, both of which are then passed through fully connected dense layers and transformed into a low dimensional latent space. Our demonstration below uses 5 input parameters and 52$\times$52 pixel input images for a total input dimension of 2709, which is then squeezed down to a 100 dimensional latent space vector which is adaptively tuned based on feedback.}
  \label{fig:TVML}
\end{figure}

\subsection{Overview of Main Results}\label{sec:main}
The general problem of time-varying distribution shift is not solvable, some assumptions are required. Consider a general nonlinear time-varying dynamic system over an interval of time $[t,t+T], T>0$:
\begin{equation}
	\dot{x}(t) = F(x(t),t), \quad x(t+T)=x(t)+\int_{t}^{t+T}F(x(\tau),\tau)d\tau, \quad y(t+T)=G(x(t+T)),
\end{equation}
where the function governing the system dynamics, $F(x,t)$, and the output measurement function $G(x)$ are analytically unknown, and $F(x,t)$ and the initial condition $x(t)$ are also time-varying. Suppose that pairs of inputs and outputs of such a system can be sampled over time
\begin{equation}
	S = \left \{(x_1,y_1), \dots, (x_n,y_n) \right \} =  \left \{ (x(t_1),G(x(t_1+T))), \dots, (x(t_n),G(x(t_n + T))) \right \},
\end{equation}
which are governed by a time-varying probability distribution $Pr(x,y,t)$. If the changes in initial condition $x(t)$ or dynamics $F(x,t)$ are arbitrarily large or fast or if the measurement function $G(x)$ is not one-to-one, it is of impossible to accurately predict $Pr(x,y,t)$ based on a finite set of data. Therefore we assume that the initial conditions of the state $x(t)$ are bounded within a compact set $\mathcal{X}$, that $F(x,t)$ is piecewise continuous in $t$ and satisfies the Lipschitz condition for some $L>0$ over $\mathcal{X}$
\begin{equation}
	\| F(x_1,t) - F(x_2,t) \| \leq L \| x_1 - x_2 \| \quad \forall \ x_1,x_2 \in \mathcal{X}, \quad \forall \ t, \label{lip}
\end{equation}
and that the variation of the system dynamics is bounded so there exists $M>0$ such that
\begin{equation}
	\| F(x_1,t_1) - F(x_2,t_2) \| < M, \quad \forall \ x_1(t),x_2(t) \in \mathcal{X}, \quad \forall \ t_1,t_2 \in [t,t+T]. \label{bound}
\end{equation} 
Condition (\ref{lip}) guarantees that each initial condition $x(t)$ has unique solution $\left \{ x(\tau), \ \tau \in [t,t+T] \right \}$ and (\ref{bound}) guarantees that all trajectories remain within a compact set \cite{khalil2002nonlinear} as trajectories satisfy bounds
\begin{equation}
	\| x_1 - x_2 \| \leq \|x_1(t) - x_2(t) \| e^{LT} + (M/L)e^{LT-1}.
\end{equation}
We also assume that the measurement $G(x)$ is invertible so that output measurement $y_i$ corresponds to a unique input $x_i$, we also assume the time-rate of change of $\|dF(x,t)/dt\| < M_F$ is bounded.

Many large complex systems have the characteristics described above, with slowly changing time-varying dynamics and time-varying bounded initial conditions. For example, as studied in detail in Section \ref{sec:particles}, large particle accelerators have complex sources of charged particle beams whose phase space distributions drift unpredictably with time and can only be measured destructively during dedicated studies. Once a machine is running to perform experiments new detailed beam measurements are typically not available, especially not of the initial beam conditions which rely on destructive methods that would interrupt operations. Furthermore the beams are accelerated and focused by groups of hundreds-thousands of interacting magnets and resonant electromagnetic field structures whose characteristics are uncertain due to misalignments and drift with time due to disturbances such as temperature drifts and vibrations. Because of uncertainties and time-variation the accuracy of simulations is limited and they require constant adaptive re-tuning \cite{scheinker2015adaptive}. Furthermore, even in the case of a time-invariant system accurate charged particle dynamics models are very slow and ML-based surrogate models which are orders of magnitude faster would be of great benefit. 

In this work we assume that we are able to perform initial training based on measurements of both system inputs and outputs, but that afterwards we must rely only on limited diagnostics and so re-training is impossible. Our goal is to adaptively tune the ML model in real-time to keep up with the time-variation of the system and of its initial conditions based only on limited output measurements. We present a method for adaptive machine learning (AML) which does not require re-training. Our approach is to incorporate a large number of physics constraints into the system so that a trained deep encoder-decoder learns the correlations between various parameters in a complex system. The case we focus on is the 6 dimensional phase space density $\rho(x,y,z,x',y',E)$ of charged particle beams where $(x,y,z)$ are particle positions, $(x',y')$ are angles of motion relative to the acceleration z-axis, and $E$ is total particle energy. We design an encoder-decoder style generative network with a 2709 dimensional input consisting of a $52\times52$ input beam $(x,y)$ distribution image as well as 5 accelerator parameter values. The output of our network is a $224\times224\times15$ pixel object which can be thought of as an image with 15 channels. Each of the 15 channels represent a 2D projection of the 6D phase space: $(x,y)$, $(x,z)$, $(x,x')$, $(x,y')$, $(x,E)$, $(x',y)$, $(x',z)$, $(x',y')$, $(x',E)$, $(y,z)$, $(y,y')$, $(y,E)$, $(y',z)$, $(y',E)$, $(z,E)$. We then demonstrate that the network has learned the correlations in the system by using measurements of only $(z,E)$, which are typically available online, to predict the $(x,x')$ and $(y,y')$ distributions, which are not easily measured in accelerators in real-time. Our approach is to adaptively tune the low-dimensional latent space of the deep encoder-decoder style CNN based on real-time feedback to quickly compensate for unknown and fast distribution shifts as shown in Figure \ref{fig:TVML}. We demonstrate our approach for predicting the properties of a time-varying charged particle beam in a compact particle accelerator. By squeezing our 2709 dimensional input space down to a general nonlinear representation in a much smaller latent space (100 in this case), we can quickly adaptively tune our system by utilizing the encoder-decoder representation learned by the CNN. This flexibility allows us to quickly respond to unknown disturbances and changes, such as unknown changes of the input images and the input parameters, which create a difference between a function of the CNN's generated predictions and some related measurement. 

This general approach can be applied to a wide range of neural network structures, including 3D CNNS, and any adaptive feedback algorithm can be used for the adaptive feedback branch of the setup shown in Figure \ref{fig:TVML} such as stochastic gradient descent or robust conjugate search. In our case we prefer to use a recently developed robust extremum seeking (ES) feedback control algorithm which was originally designed for the stabilization of unknown open-loop unstable time-varying nonlinear systems and then extended for the optimization of many parameter time-varying unknown systems with noise-corrupted measurement functions \cite{scheinker2012minimum,scheinker2016bounded}. This method can quickly tune many coupled parameters and has been applied for time-varying problems such as adaptively learning optimal feedback control policies for unknown time-varying systems directly from measurement data \cite{scheinker2020extremum}.

\section{Latent Space Tuning}

\subsection{Machine Learning, Dynamic Systems, and Control}
Our approach of combining external feedback with neural networks is analogous to a natural phenomenon in biological systems in which networks of neurons interact with each other and are controlled by external feedback loops and other networks for tasks such as synchronization in the presence of delays, noise, and disturbances \cite{doho2020transition,ibrahim2019complex,ibrahim2021lag,nobukawa2019resonance,nobukawa2019controlling}. Some initial AML studies have coupled the outputs of CNNs to adaptive feedback for real-time accelerator phase space control \cite{ref_ES_ML_FEL} and for predicting 3D electron density distributions for 3D coherent diffraction imaging \cite{ref_ES_ML_CDI}. Powerful ML methods have also recently been developed for control applications, for describing dynamic systems using advanced methods such as Koopman theory and dynamic mode decomposition, and for learning the partial differential equations that govern complex physical systems directly from data \cite{tu2013dynamic,brunton2016discovering,rudy2017data,lusch2018deep,brunton2019data,brunton2020machine}.

\subsection{Adaptive Latent Space Tuning}
The approach presented here is applicable to any type of neural network (such as 3D CNNs), we focus on convolutional encoder-decoder architectures as they are incredibly powerful for working with 2D distributions (images) directly. The inputs to our ML structure can be split into two categories: one set $(\mathbf{v}_p,I^0)$ is for training the CNN and consists of a vector of scalar parameters $\mathbf{v}_p$ of length $N_p$ which represent system components such as magnets or electric fields in our example and an $N\times N$ input image $I^0$ which in the example below represents the initial input beam $\rho(x,y)$ density distribution entering the accelerator. The second set of inputs, $\mathbf{v}_{L,c}$, is a vector of adaptively tuned control parameters of length $N_L$ which are kept equal to 0 during training, which enter directly at the low dimensional latent space at the bottleneck of the network, as shown in Figure \ref{fig:TVML}. In terms of training, our input feature space has total dimension $N_{i,p} = N_p + N^2$ and $N_L$ is chosen such that $N_L \ll N_{i,p}$ with the network designed to pinch down to a minimal latent space dimension of $N_L$. 

When a $N\times N$ input image represented by matrix
\begin{equation}
    I^0=\left \{ I^0 \left (i_{0},j_{0} \right ), i_{0},j_{0} \in \left \{1,2,\dots,N \right \} \right \},
\end{equation}
passes through a stride 2 convolutional layer with a $3\times 3$ filter $F_{0,ij}$ the output image size is reduced by a factor of 4 resulting in a $N/2\times N/2$ image $I^1$ whose pixels are defined as
\begin{eqnarray}
    && I^1_{i_{1},j_{1}} = f \left (b^0 + \sum_{i=-1}^{1}\sum_{j=-1}^{1}F_{0,ij}\times I^0_{i_{0}+i,j_{0}+j} \right ), \\
    && i_{0},j_{0} \in \left \{1,3,\dots,\hat{N} \right \}, \quad i_{1},j_{1} \in \left \{1,2,\dots,N/2\right \},
\end{eqnarray}
where $b^0$ is the bias and $f$ a nonlinear activation function, $\hat{N}$ is equal to $N-1$ if $N$ is even and $N$ if $N$ is odd, with extra zeros padded onto the image as needed. Typically a collection of $N_f>1$ filters is used, which even for a single layer CNN would result in an output image of the form
\begin{equation}
    I^1_{i_{1},j_{1}} = b^1 + \sum_{n=1}^{N_f} w_n \times f_n \left (b^0_n + \sum_{i=-1}^{1}\sum_{j=-1}^{1}F_{0,ij,n}\times I^0_{i_{0}+i,j_{0}+j} \right ), \label{CNN}
\end{equation}
with $11\times N_f + 1$ total adjustable weights and biases for just a single layer CNN. For a wide encoder-decoder CNN with several layers the total number of adjustable parameters easily grows to millions and re-training requires large collections of new data sets. In our approach, as is typically the case with encoder-decoder CNNs, after several layers of convolutions we significantly reduce the size of a relatively large image ($52\times 52 \rightarrow 7\times 7 $) using more and more filters at each stage, ending up with a tensor of shape $N_i\times N_i \times N_f$ where $N_i \times N_i$ is the final image size and $N_f$ is the number of filters in the last convolution layer of the encoder. We flatten the image and apply dense fully connected layers to reach a low-dimensional latent space representation, a vector $\mathbf{v}_L$ of length $N_L \ll N_{i,p}$. We then add an $N_L$ dimensional control input vector $\mathbf{v}_{L,c}$ so that the latent space parameters are given by
\begin{equation}
    \mathbf{p}_{L} = \left ( p_1,\dots,p_{N_L} \right) = \left ( v_1  + v_{c,1}, \dots, v_{N_L}  + v_{c,N_L} \right ) = \mathbf{v}_L + \mathbf{v}_{L,c},
\end{equation}
where the vector $\mathbf{v}_L$ is the output of the trained encoder and $\mathbf{v}_{L,c}$ the controlled parameters used for adaptively tuning the latent space ($\mathbf{v}_{L,c}=0$ when training). The vector $\mathbf{p}_L$ is passed through additional fully connected dense layers before being reshaped into a small image ($\sim8\times 8$) which then passes through a series of 2D transpose convolution layers until a collection of $N_o$ output images of size $N_{im}\times N_{im}$ are generated in a final layer with $N_c$ channels with size $\hat{I}(i,j,d) = N_{im}\times N_{im} \times N_c$. Once the network is trained this collection of output images is a general nonlinear function (the generative branch of the network) of the parameters $\mathbf{p}_L$ of the form
\begin{equation}
    \hat{I}(i,j,d) = \mathbf{F}\left ( \mathbf{p}_L,\mathbf{w},\mathbf{b}, \left \{ A \right \} \right ), \quad i,j \in \left \{1,2,\dots,N_o \right \}, \quad d \in \left \{1,2,\dots,N_c \right \},
\end{equation}
where $\mathbf{w}$ and $\mathbf{b}$ are the weights and biases and $\left \{ A \right \}$ are the set of activation functions of the generative layers. Our prediction $\hat{I}$ is our estimate of some unknown physical quantity $I$ which we assume we cannot easily directly measure fully. In order to enable the adaptive feedback part of this procedure we must assume that we have some form of non-invasive online measurement of $M(I(t))$ that can be compared to a simulated measurement of our prediction $\hat{M}(\hat{I})$, which we know how to approximate. For example we may be interested in a 6D density distribution but we only have direct measurements of one 2D projection. A detailed example and simulation study of such a problem for particle accelerator applications is presented in Section \ref{sec:particles}. We set up a dynamic feedback loop for minimization of a cost function of the form
\begin{eqnarray}
    C(\mathbf{p}_L(t),M(I(t))) &=& \mu\left (\hat{M}\left (\hat{I} \left (\mathbf{p}_L(t) \right ) \right ),M(I(t)) \right ), \label{cost} \\
    \frac{\partial p_i}{\partial t} &=& \frac{\partial v_{c,i}}{\partial t} = \sqrt{\alpha\omega_i}\cos \left ( \omega_i t + k C(\mathbf{p}_L(t),M(I(t))) \right ), \label{ES}
\end{eqnarray}
where $\mu$ is a metric defined to quantify error, such as mean squared error between 2D images:
\begin{equation}
    \mu\left (M(i,j),\hat{M}(i,j)\right ) = \frac{1}{N_xN_y}\sum_{i=1}^{N_x}\sum_{j=1}^{N_y}\left ( M(i,j) - \hat{M}(i,j) \right )^2. \label{mse}
\end{equation}
The adaptive feedback dynamics (\ref{ES}) are chosen based on the results in \cite{scheinker2012minimum,scheinker2016bounded,scheinker2020extremum}. The hyper-parameters in (\ref{ES}) can intuitively be understood as $\alpha$ representing a dithering amplitude which controls the size of the dynamic perturbations, $k$ a feedback gain, and the product $k\alpha$ can be thought of as an overall learning rate. The dithering frequencies $\omega_i$ are chosen relative to a base frequency $\omega$ such that they are distinct, of the form $\omega_i = r_i \omega \neq r_j \omega = \omega_j$ for $i\neq j$ such that no two frequencies are integer multiple of each other (such as distinct $r_i \in [1,1.75]$) because non-linearity typically introduce harmonics into the system dynamics. The convergence results for this feedback algorithm depend on the parameters being orthogonal in Hilbert space such that for any $t>0$ and any measurable $f(t) \in L^2[0,t]$, the $L^2[0,t]$ inner products in the limit of large frequency $\omega$ are:
\begin{equation}
    \lim_{\omega \rightarrow \infty} \int_{0}^{t}\cos(\omega_i\tau)f(\tau)\cos(\omega_j\tau)d\tau = 0,  \quad \lim_{\omega \rightarrow \infty} \int_{0}^{t}\cos^2(\omega_i\tau)f(\tau)d\tau = \frac{1}{2} \int_{0}^{t}f(\tau)d\tau.
\end{equation}
In fact any orthogonal functions can be used, including non-differentiable and discontinuous square waves, as described in more detail in \cite{scheinker2012minimum,scheinker2016bounded,scheinker2020extremum}. The resulting on average dynamics of the evolution of the cost function with the latent space variables evolving under feedback (\ref{ES}) are then given by
\begin{eqnarray}
	\frac{d C}{dt} &=&   \nabla^T_{M}C\frac{\partial M}{\partial t}  + \left (\nabla_{\mathbf{p}_{L,c}}C\right )^T\frac{\partial \mathbf{p}_{L}}{\partial t} =  \nabla^T_{M}C\frac{\partial M}{\partial t}  - \frac{k\alpha}{2}\left ( \nabla_{\mathbf{p}_{L,c}}C \right )^T\left (\nabla_{\mathbf{p}_{L,c}}C \right ),
\end{eqnarray}
which, based on the assumptions made in section \ref{sec:main}, in the case of the cost function being convex and positive semi-definite so that it can be considered as a Lyapunov function for the overall system dynamics (\ref{cost}), (\ref{ES}), result in the latent space parameters tracking a unique global minimum when the feedback gain $k\alpha>0$ is sufficiently large relative to the time variation of the system $\nabla^T_{M}C\frac{\partial M}{\partial t}$ which is bounded over compact sets by the assumptions above, such that $dC/dt<0$. In the case that the cost function is not convex we may still maintain a globally optimal set of latent space parameters as long as we started there and track it with time, otherwise it is possible to settle in a local minimum. In practice, if starting with a good training set, the system is able to uniquely track the time-varying global optimal with time based only on limited output measurements, as demonstrated below.

\section{Particle Accelerator Application}\label{sec:particles}

\begin{figure}
  \centering
  \includegraphics[width=0.6\textwidth]{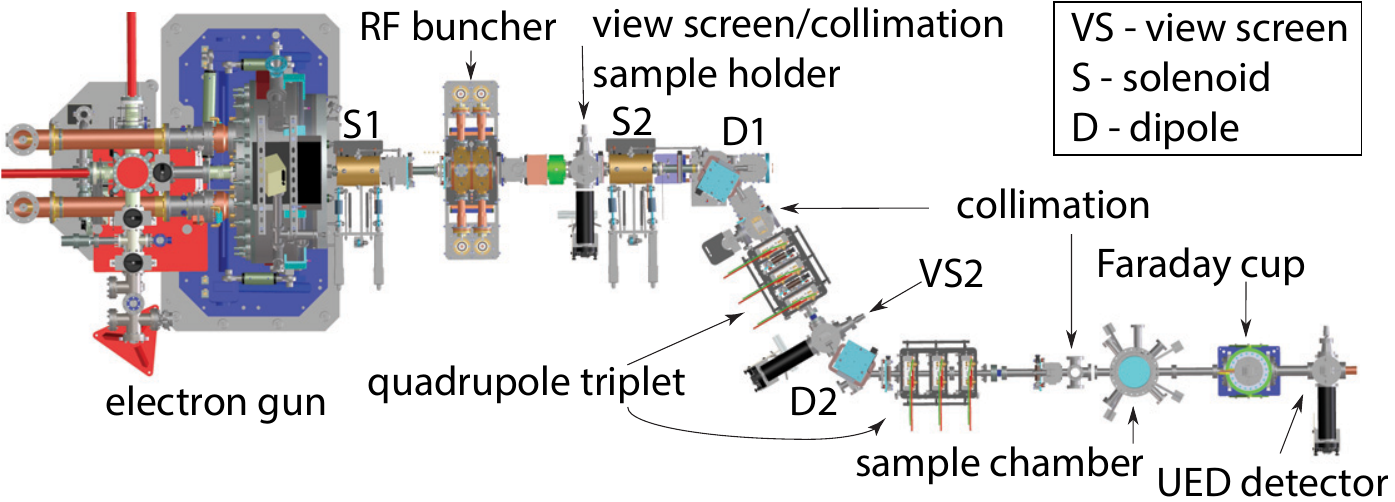}
  \caption{The HiRES ultrafast electron diffraction beamline is shown (adapted from \cite{ref_HiRES}). For our work we care about the fact that the input beam distribution at the electron gun varies with time and the characteristics of the accelerator itself are also not perfectly reproducible including the magnetic fields of the solenoids and quadrupole magnets.}
  \label{fig:HiRES}
\end{figure}

Charged particle bunches in particle accelerators can be composed out of anywhere from $10^5$ to $10^{10}$ particles depending on the total charge. The charged particle dynamics are defined over a 6D phase space with density $\rho(x,y,z,x',y',E)$ where $(x,y,z)$ are the physical particle locations with $z$ typically being the axis of acceleration and $(x,y)$ the transverse off-axis distance. The components $(x',y') =  (p_x/p_z,p_y/p_z)$ are the angles of particle motion relative to the accelerator axis z, where the $(p_x,p_y,p_z)$ are components of particle momentum and $E$ is total particle energy. All 6 dimensions are coupled through collective effects such as space charge forces which can simply be described as the electric fields of all of the individual particles pushing each other apart and via coherent synchrotron radiation in which accelerating charged particles release light which impacts other particles in the bunch thereby changing their energy \cite{ref_LL}. The influence of collective effects grow as accelerators generate shorter more intense bunches such as 30 fs bunches at the SwissFEL X-ray FEL \cite{ref_Malyzhenkov_attosecond}, sub 100 fs bunches for ultra fast electron diffraction (UED) \cite{ref_compression}, and picosecond bunch trains for UEDs and multicolor XFELs \cite{ref_microbunching}. Our work focuses on the High Repetition-rate Electron Scattering apparatus (HiRES, shown in Fig. \ref{fig:HiRES}) at Lawrence Berkeley National Laboratory (LBNL), which accelerates pC-class, sub-picosecond long electron bunches up to one million times a second (MHz), providing some of the most dense 6D phase space among accelerators at unique repetition rates, making it an ideal test bed for advanced algorithm development \cite{ref_HiRES,ref_HiRES_UED}. 

In order to better understand and compensate for collective effects in intense beams it would be very helpful to have a view of the full 6D phase space in real time, which is of course impossible. What is usually possible is to record various 2D projections of a beam's 6D phase space. For example, it is possible to measure the longitudinal phase space (LPS) of a charged particle bunch by using a transverse deflecting radio frequency resonant cavity (TCAV) which measures $(z,E)$. 

\begin{figure}
  \centering
  \includegraphics[width=1.0\textwidth]{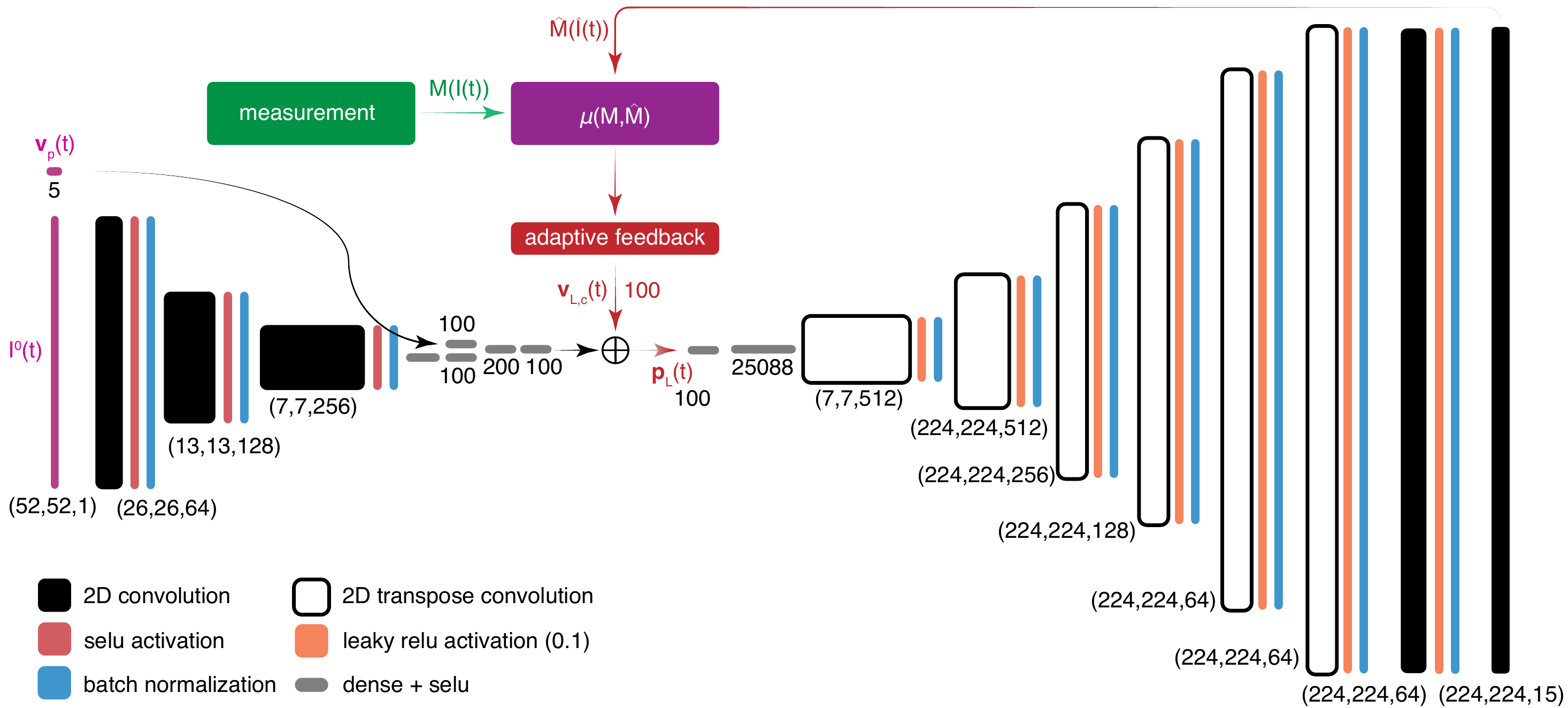}
  \caption{The encoder-decoder CNN structure for the AML setup is shown with layer sizes such as $(224,224,15)$ representing an output of 15 filters of image size $214\times214$ each. The dense layer widths shown as single numbers.}
  \label{fig:AML_GAN}
\end{figure}

Although physics and machine learning-based surrogate models can potentially serve as non-invasive beam diagnostics, the main challenges they face are uncertainty in and time-variation of accelerator components, such as magnets and accelerating resonant cavities, and the time-variation of the initial beam distribution entering the accelerator. Such distributions drift with time requiring lengthy measurements that interrupt accelerator operations. Approaches to adaptively tune online physics models have been developed which attempted to track time-varying beam properties based on non-invasive measurements \cite{scheinker2015adaptive}, but are limited by the slow speed of physics models (several seconds per simulation) and require accurate knowledge of the time varying input beam distribution.

\begin{figure}
  \centering
  \includegraphics[width=0.9\textwidth]{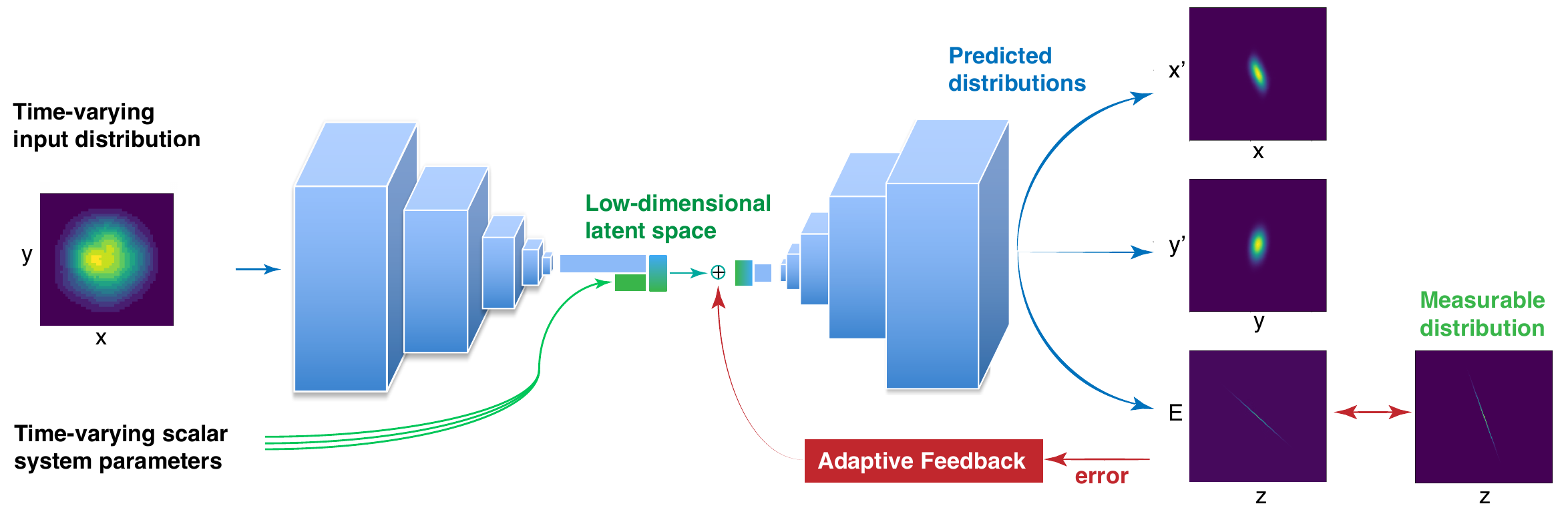}
  \caption{Adaptive latent space tuning setup for predicting the transverse phase space of the HiRES beam based only on longitudinal phase space measurements.}
  \label{fig:AMLES}
\end{figure}

\begin{figure}
  \centering
  \includegraphics[width=0.7\textwidth]{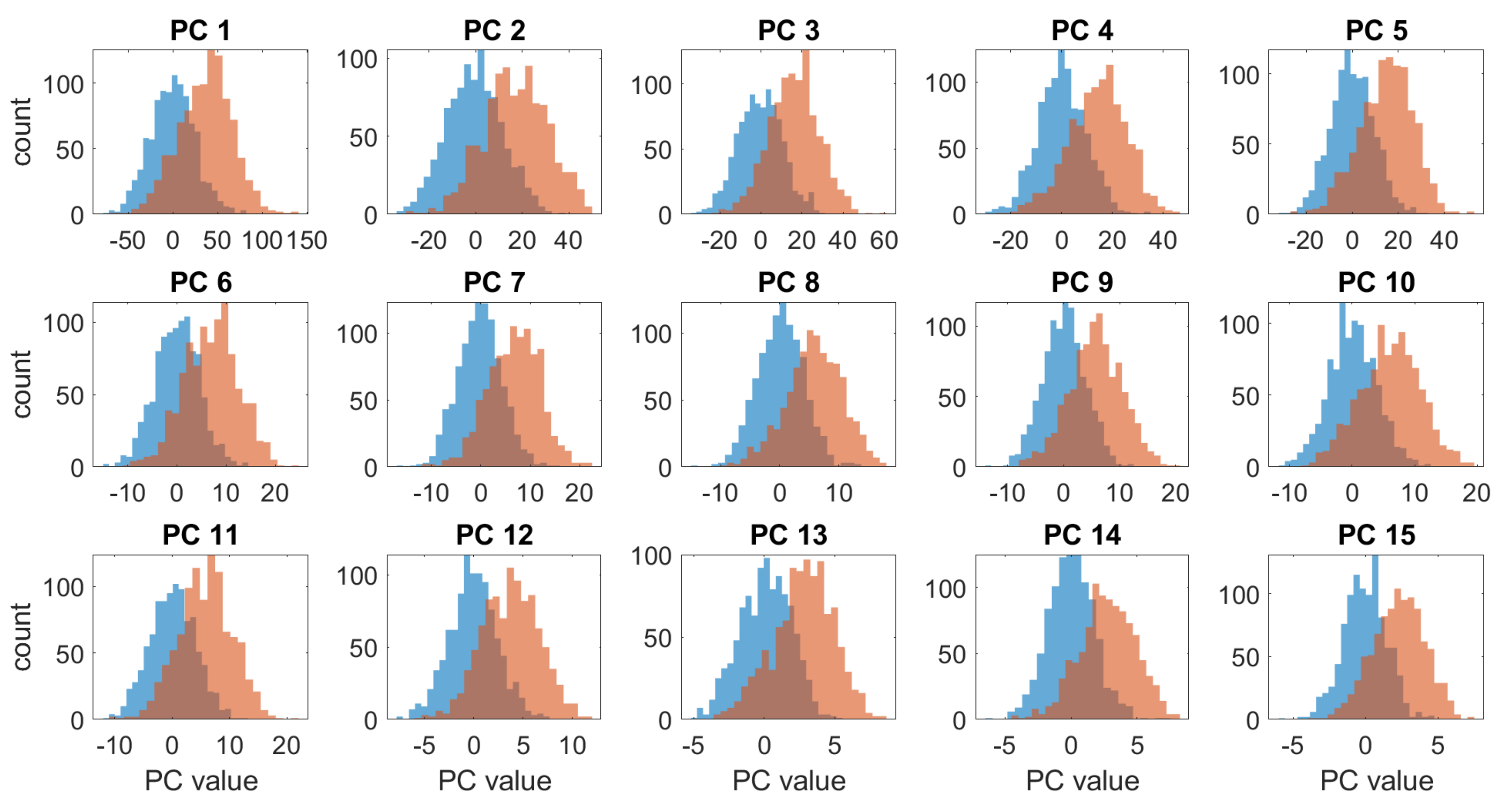}
  \caption{Histograms of 1000 samples of the first 15 PCA components of the training set input beam distribution (blue) and the new input beam distribution (red).}
  \label{fig:dist_shift}
\end{figure}

\begin{figure}
  \centering
  \includegraphics[width=1.0\textwidth]{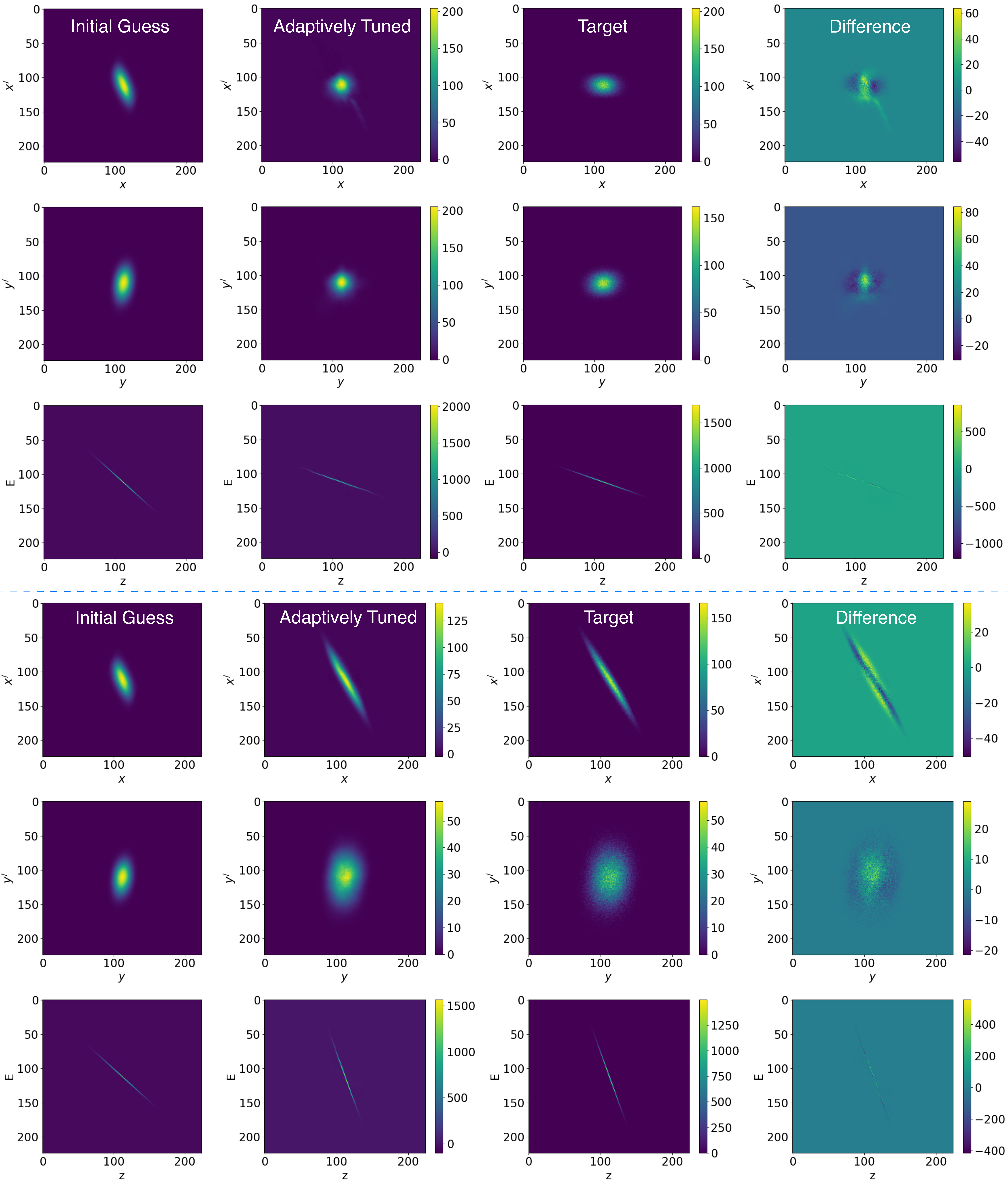}
  \caption{Two results of adaptively tuning the latent space based only on comparisons of the $(z,E)$ 2D phase space projections resulting in a close match of all 3 phase space projections.}
  \label{fig:tuning_results_1}
\end{figure}

We propose to utilize AML to develop a virtual diagnostic that is able to provide information about the beam's phase space by using only available 2D measurements of the LPS $(z,E)$ by learning the physical relationships and correlations between the various phase space slices for a particular accelerator. Because the 6D phase space dynamics are coupled and the charged particles evolve under physics constraints unique to a given accelerator lattice, it may be possible to uniquely recover or track various 2D slices of the 6D phase space based on just a single 2D measurement. The 6D phase space distribution $\rho(x,y,z,p_x,p_y,p_z)$ approximately satisfies the relativistic Vlasov equation
\begin{eqnarray}
    && \frac{\partial \rho}{\partial t} + \mathbf{v} \cdot \mathbf{\nabla}_\mathbf{x}\rho + \dot{\mathbf{p}}\cdot \mathbf{\nabla}_\mathbf{p}\rho=0, \quad \mathbf{v} = (\dot{x},\dot{y},\dot{z}), \quad \mathbf{p} = (p_x,p_y,p_z) \\
    && \dot{\mathbf{p}} = q\left ( \mathbf{E} + \mathbf{v} \times \mathbf{B} \right ), \quad \mathbf{p}=\gamma m \mathbf{v}, \quad \gamma = 1/\sqrt{1-\frac{v^2}{c^2}}, \quad v = |\mathbf{v}|,
\end{eqnarray}
where the electric and magnetic fields include contributions from charges within the beam as well the external electromagnetic fields of accelerator components such as radio frequency (RF) resonant accelerating cavities, solenoids, dipoles, and quadrupole magnets. To learn the correlations within the HiRES electron beam we simulated the evolution of 210 thousand sets of various initial beam distributions and accelerator parameter settings $\mathbf{p}=(p_1,\dots,p_5)$ which were the beam energy at the accelerator entrance, the phase of the RF electromagnetic field in the gun generating the beam, the peak field in the buncher cavity directly after the gun, the buncher RF field phase, and the solenoid magnet's current. For each simulation we recorded the randomized initial $52\times52$ pixel $(x,y)$ beam distribution and all 15 2D slice combinations of the final 6D beam distribution down stream as 15 $224\times224$ pixel images. An encoder-decoder type CNN was then trained using the input $(x,y)$ beam distribution and accelerator parameters $\mathbf{p}$ as inputs. Back propagation was carried out using the Adam optimizer which trained on the entire data set with a batch size of 10 with a learning rate of $10^{-3}$ the first 3 passes through the data, then trained again with a learning rate of $10^{-4}$ on another 3 passes, and then finally with $10^{-5}$ on a single NVIDIA Quadro RTX-5000 GPU. All weights within the layers were initialized with their default settings. We then tested if we could recover the $\rho_{x,x'}(x,x')$ and $\rho_{y,y'}(y,y')$ transverse 2D phase space distributions by adaptively matching the network's prediction of the 2D longitudinal phase space distribution $\rho_{z,E}(z,E)$ as shown in Figure \ref{fig:AMLES}, where the 2D distributions are projections of the 6D phase space:
\begin{equation}
    \rho_{z,E}(z,E) = \int_x\int_{x'}\int_y\int_y' \rho(x,y,z,x',y',E)dxdx'dydy.
\end{equation}
To demonstrate the approach we generated a new set of 1000 input beam distribution and input parameter values that were outside of the range of values used to train as shown in Figure \ref{fig:dist_shift}, for which we do net expect the network's prediction to be accurate. Furthermore, we used an input image and input parameters which were generated as the averages of the training data because we are assuming that in practical applications we will not have accurate measurements of the inputs, the results are shown in the left "initial guess" column of Figure \ref{fig:tuning_results_1}. Adaptive feedback was based on comparing the predicted and measured $(z,E)$ projections based on error defined in Equation (\ref{mse}) which was minimized by iteratively tuning the 100 parameters $\mathbf{v}_{L,c}=(v_{c,1},\dots,v_{c,100})$ of the low dimensional latent space, providing estimates of the transverse phase space $(x,x')$, $(y,y')$ as shown in Figure \ref{fig:tuning_results_1}. The distribution and accelerator parameters used in Figure \ref{fig:tuning_results_1} were closer to the training set and resulted in a very close match between all 3 phase space projections. For a distribution and parameter settings further away from the training set, as expected the matches were worse, but still impressive as shown in the bottom of Figure \ref{fig:tuning_results_1}. A degradation in predictive power as we move further from the span of the training set is expected because even with the learned relationships between the phase space projections and adaptive tuning there is no guarantee that a match of the $(z,E)$ distributions will result in a unique reconstruction of the associated transverse phase space.

\section{Conclusions}
We have demonstrated a new approach to adaptive machine learning for time-varying systems by adaptively tuning the low dimensional latent space representation of complex systems. In our demonstrative example, by utilizing an encoder-decoder architecture we were able to reduce an input (image + 5 inputs) of dimension 2709 down to a 100, which allowed us to tune in real-time without relying on re-training which would be much slower or in our case impossible because we lose access to input beam distribution measurements during operations. Our approach taught the encoder-decoder the relationships between the various 2D projections of a 6D phase space for the HiRES compact accelerator, which allowed us to predict and track the transverse phase space based only on longitudinal phase space measurements, no longer requiring accurate accelerator parameter or input distribution measurements. Although we focused on a particle accelerator-based demonstration and it is outside of the scope of this short conference paper to study a wider range of systems, we believe that this approach can be used for general machine learning problems where limited data available for making predictions after a network has been trained to learn underlying relationships and for problems with time-varying non-stationary distributions. Furthermore, this approach will be useful for a wide range of advanced particle accelerators with online real-time $(z,E)$ longitudinal phase space diagnostics such as FACET-II, LCLS, LCLS-II, EuXFEL, and SwissFEL. 

{
\small

\bibliography{Adaptive_Latent_Space_Tuning_v9}

}

\end{document}